\documentclass[runningheads]{llncs}

\usepackage[T1]{fontenc}
\usepackage{graphicx}
\begin{document}
\title{Improving AGI Evaluation: A Data Science Perspective}
%

\author{John Hawkins}

\authorrunning{J. Hawkins.}
%
\institute{}
\institute{Pingla Institute, Sydney, Australia \\
email{john@getting-data-science-done.com}\\
\url{http://www.pingla.org.au}}

\maketitle              
\begin{abstract}
Evaluation of potential AGI systems and methods is difficult due to the
breadth of the engineering goal. We have no methods for perfect evaluation of
the end state, and instead measure performance on small tests designed to 
provide directional indication that we are approaching AGI. In this work we argue that
AGI evaluation methods have been dominated by a design philosophy that 
uses our intuitions of what intelligence is to create synthetic tasks, that
 have performed poorly in the history of AI. 
Instead we argue for an alternative design philosophy focused on 
evaluating robust task execution that seeks to demonstrate AGI through 
competence. This perspective is developed from common practices in data science
that are used to show that a system can be reliably deployed. We provide practical
examples of what this would mean for AGI evaluation.

\end{abstract}
\section{Introduction}

The field of Artificial General Intelligence (AGI) is demarcated by an explicit desire to
develop systems exhibiting a stronger form of intelligence than the highly successful
narrow AI systems that are routinely used by data scientists and machine learning
engineers. Multiple broad definitions exist for what the core of \textit{intelligence} or 
\textit{general intelligence} involves\cite{10.5555/1565455.1565458}. 
Within the field of AGI the definitions typically invoke notions of achieving 
multiple goals in different contexts\cite{goertzel2014artificial,10.5555/1565455.1565458}, 
while dealing with changing, novel and ambiguous environments\cite{wang2019defining},
exhibiting autonomy and an ability to learn from experiences\cite{voss2007}.
These definitions are sufficient to drive research interest, but it has been argued that 
they are ambiguous enough that they prevent definition of robust and universal evaluation methods
for the field to focus on\cite{Wang2010/06}. Many of the metrics that do exist are
prone to the long standing problem that they can be gamed through efforts to circumvent 
the intentions behind the metric
and create a system that in some senses tricks us\cite{Bringsjord2001}.
Furthermore, critics of the field have 
argued that the lack of strong definitions and metrics for AGI have
rendered it not just unachievable, but also the source
of serious misallocation of concerns about AI risk \cite{Mueller2024}.

Previous meta-analysis has suggested that evaluation
methods can be sorted into empirical and theoretical approaches\cite{Wang2010/06}, 
where empirical here means evaluated in terms on similarity to human intelligence,
and theoretical means evaluation in agreement with a theoretical construct of 
intelligence. However, as these empirical evaluations involve choosing datasets
that reflect human intelligence, they are always guided by some theoretical notion of 
intelligence\cite{Wang2010/06}.

In spite of the wide variety and ambiguity of definitions, there are commonalities between
definitions of what general intelligence would be and how it should be measured. 
Generality and flexibility feature highly in most definitions, with the core differences
emerging from defining how you concretely determine how to measure these qualities.
In addition, the emerging focus on the notion of an AGI being able to replace a human
for any productive task\cite{Nilsson_2005}, has led to an emphasis on defining the level
of agency and autonomy in a system\cite{10.5555/3692070.3693548}. 
Autonomy is required because a system that needs to be micro-managed
through all tasks offers very little in the way of economic advantage, and if it is
unable to perform without the assistance of human cognition it suggests that it is not
a complete cognitive entity. 

If our empirical evaluations of AGI are guided by one of more theoretical structures,
it is worth asking how well these theories are working. 
Multiple authors have observed that the poverty of our AI benchmarks results
in a constant adjustment of whether a given task requires intelligence or not
\cite{McCorduck2004}. This may be in part because our benchmark driven evaluation means
that we place emphasis on performance alone, rather than the methodology
used by an algorithm to achieve results\cite{chollet2019measureintelligence}.
Which is further supported by the trend over time in which we re-evaluate our
evaluation methods in part by understanding the mechanisms by which machine learning
systems seem to game them. This observation supports the empirical/theoretical division
in the field\cite{Wang2010/06}. Certain researchers want to advance progress through better
empirical task based evaluations, while others want evaluations that provide insight
into the emerging cognitive architectures that produce results through learning.
Among the modern empirical evaluations there is a heavy emphasis on both 
variety\cite{Adams_Banavar_Campbell_2016,zhong2024},
and abstract reasoning\cite{chollet2019measureintelligence}. 
It is worth noting that the Large Language Models (LLMs) that are being touted as a 
potential form of AGI\cite{norvig2023}, are routinely evaluated by a
wide variety of criteria\cite{guo2023evaluatinglargelanguagemodels,10.1145/3641289},
in which reasoning capability is only one part.
Furthermore, there are a wide range of potential problems with
these models beyond limitations of reasoning ability\cite{kaddour2023challengesapplicationslargelanguage}

We propose that a central tension in the evaluation of AGI is the desire for agents that
can perform arbitrary human jobs, but are then evaluated on the basis of general performance
across narrow individual tasks. Recognition of this tension is implicit in proposals
like Wozniak's "Cup of Coffee"\cite{Wozniak}, such as an ability to prepare
meals in arbitrary kitchen\cite{marcus2022}, the "Employment Test" in which an entire
human job is completed\cite{Nilsson_2005}, 
or exhibiting self-directed behaviour in a physical environment\cite{peng2024}.
If we remove the requirement for physical embodiment,
in line with the proposal that evaluation should focus on cognitive and meta-cognitive 
tasks\cite{10.5555/3692070.3693548}, then we are looking for methods that evaluate models
using performance on completely digital knowledge works\cite{Nilsson_2005}.

Data science has emerged as the field in 
which people versed in statistics and machine learning apply their technical skills
to generate systems that are intergrated in business processes. 
In the process, the practice of data science
has developed many methods of evaluation designed to ensure that their systems perform
reliably in the world. In this work we draw on the rigorous evaluation methods of
data science as a source of inspiration for pragmatic evaluation for AGI systems.

The methods presented can be summarised as empirical evaluations designed to mitigate
the problem of data set biases, contamination and potential memorisation. 
Models that simply memorise training data pose
risks of security, but also impede accurate theoretical understanding of what the learning
process is achieving. These problems exist for machine learning models in 
general\cite{feldman2020,ye2024}, as well as the new generative foundation models 
being touted as some form of AGI\cite{bubeck2023sparksartificialgeneralintelligence,norvig2023}.
For our purpose, mitigating potential memorisation effects is essential so that any 
evaluations of an AGI demonstrate a model can reliably perform work outside the
bounds of its training data.

\subsection{Theories of Empirical Evaluation}

Early methods for evaluating whether a system possessed human-like intelligence, 
focused singular activities with varying degrees of complexity, from 
performing well at perfect information games\cite{Levinson1991},
to conducting a conversation well enough to appear human\cite{Akman2000}.  
Modern AI research arguably began once we had a set of machine learning algorithms
that could perform very well on a series of cognitive tasks intended to define unique 
qualities of human intelligence\cite{Adams_Banavar_Campbell_2016}. 
Many of these tasks are now routinely performed by AI systems, furthermore some of
them are performed by the general purpose AI systems\cite{zhong2024} 
built using variations of the Transformer architecture\cite{Vaswani2017}. 
Nevertheless, research into benchmarks and evaluation of 
AGI systems continues because these systems all fall short of many notions of
what an AGI should be capable of, in part because they are not capable of doing a complete
human job, and in part due to suspicions that their performance is largely training data
memorisation\cite{tirumala2022,hartmann2023}.

In response to these concerns more recent AGI testing frameworks have moved to include very
abstract pattern recognition problems similar to IQ testing\cite{chollet2019measureintelligence}.
This approach seeks to define reasoning tasks that are impervious to memorisation and capture the
capacity for abstraction and analogical reasoning that we suspect underlies human ability to
solve problems in novel domains. However, as pure LLM based systems make incremental progress
on these tasks\cite{pfister2025}, it again raises the prospect that these tasks may not be immune to
the problems with earlier evaluations.

An alternative approach has been to suggest that we test the ability of a proposed AGI system to thrive inside
a simulated world. These approaches attempt to remove all possibility of over-fitting to training
on pre-determined tasks, and instead seek to test the ability of the system to infer the causal
principles in the world necessary to make good decisions and survive\cite{Xu2023}. A similar proposal
suggests evaluation occurs inside constrained worlds which mimic human learning and development 
through interactive lessons\cite{Bugaj2009/06}. In such worlds evaluation occurs in tandem with
learning. The AGI system must demonstrate a capacity to learn and generalise tasks from either
the mechanics of the world, or its social interactions, thereby addressing the critique that
how an AGI learns should be part of the evaluation.

The simulated world approach will no doubt prove essential to deep theoretical insights into
how to develop the right kinds of cognitive architectures to achieve AGI. 
However, in the short term, we face a field in which significant research effort is directed
toward incremental improvements made with brute force data driven agents built using 
variations on an expanding parameterisation of the Transformer decoder\cite{pfister2025}. 
We therefore propose that robust evaluations that target the pragmatic goals of this 
segment of the field are essential.

\section{Methods}
 
\subsection{Agency}

One of the ways of differentiating between tasks that an AGI could be evaluated on are
the levels of agency required to find, or even initiate, a solution. For example,
take the idea of testing if an AGI can create a novel research paper in a given field,
given a predetermined set of reference articles.
This task is higher agency than a standard machine learning task in research assistance,
such as filtering potential references for the presence of a randomised control trial\cite{Cohen2015},
indicators of research quality\cite{Lokker2023},
or a semantic match with the research proposal\cite{10.1145/3676581.3676582}.
Nevertheless, this task is relatively constrained, the input references and the topic of
research is being given to the agent. Contrast this with the actions of the
human beings who lead research groups and guide the actions of PhD students and
post-doctoral researchers. Lead researchers need to conceive of the entire research
project by identifying the set of key references and the research gap that exists between
them. This is a high agency task directed only by their incentive to produce novel research
that can attract grant funding.

\begin{table}
\caption{Agency Calibration in Evaluation Tasks}\label{tab:agency}
\begin{tabular}{|l|l|l|}
\hline
Level             & Explanation\\
\hline
High              & Point to the general domain and ask for the system to find and solve problems\\
Medium            & Point to the specific type of problem and ask for a solution\\
Low               & Point to the specific problem and provide details of what to include in a solution\\
\hline
\end{tabular}
\end{table}

Similar to previous research we propose that evaluation 
of AGI need to include quantification of the agency required for a given task, 
such that we can determine progress
towards truly autonomous AGI agents\cite{10.5555/3692070.3693548}. 
However, we consider autonomy to be something that can be exhibited at differing degrees
even by human beings, depending on the nature of the job they are doing.
To that end we have augmented our examples by generating subtle variations of any task
with a simple notion indicating how the level of agency varies with these task variations. 
These levels are explained in Table \ref{tab:agency}.

\subsection{Data Science Perspectives} 

The primary principle in the evaluation of data science systems is that \textit{separation
of training and testing data needs to be impervious to leakage}. This principle is understood
in machine learning research, but routinely circumvented through the creation of custom data
sets to target a given benchmark\cite{pfister2025}, simple mistakes in the coding of 
experiments\cite{alomar2025} or multiple test data queries in a way that compromises
competitive machine learning challenges\cite{blum2015}.

The closer a set of data science models are to the core of a business (for example banking or
insurance), the greater scrutiny there tends to be on the robustness of the testing protocols
and the integrity of test data. To this end, data scientists have developed multiple protocols
for testing an algorithm using different test sets to understand the potential weakness of an
algorithm. In this section we demonstrate how these principles can be applied to develop
AGI testing protocols.
 
\subsubsection{Out of Time Testing}

\begin{figure}
\includegraphics[width=\textwidth]{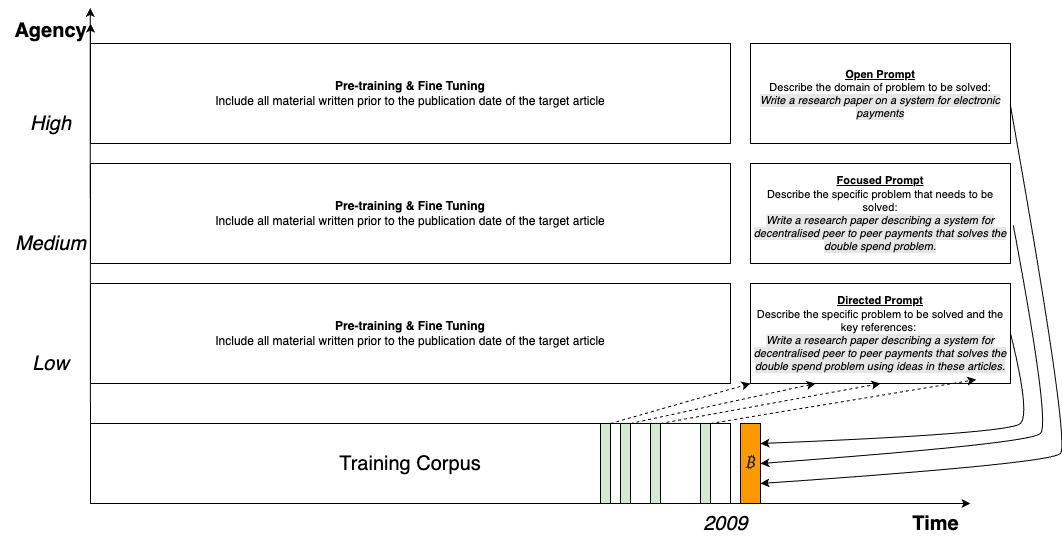}
\caption{Out of time evaluation of AGI on research tasks.}
\label{fig:outoftime}
\end{figure}

Out of time testing is a common practice in the evaluation of time series 
forecasts, involving separation of data by a point in time called the
'forecast origin' such that data points prior to this point are used 
to fit a model and points after it are used to evaluate the model
\cite{TASHMAN2000437,Hewamalage2022}. Many variations of this method have been adopted,
but the core principle involves recognising that data from the future cannot
be used when simulating model performance. The out-of-time approach has 
been gradually adopted more broadly in data science with the recognition
that non time-series problems can also have time dependent qualities 
(such as concept or data drift) that necessitate time based evaluations, 
or algorithm design\cite{10.1145/2783258.2783372,10.1007/s10489-021-02735-2}. 

When we are looking to evaluate a potential AGI system, and the algorithms are
potentially trained on vast amounts of human output, we need to be certain that the
system is not merely presenting memorised results of human cognition.
To ward against this possibility it is crucial that our evaluation metrics
ensure that training data is separated using time stamps so models
can be evaluated for their ability to recreate solutions equal, or equivalent, to
those produced by human cognition, without depending on the specific human
outputs where the problem is already solved.

We provide an example of what out-of-time testing could look like for a potential
AGI system in Figure \ref{fig:outoftime}. The diagram illustrates that to understand
if an AGI is capable of research work it should be able to create similar work to
humans with the same inputs, and critically without being trained on the outputs
of the human work. We choose the Bitcoin white-paper as the example\cite{nakamoto2009bitcoin},
as this is a piece of research that requires no experimental work in the real world, 
but requires combining the work of many different researchers to solve an unsolved problem.
The most important notion in this proposal is that all training data, pre-training
and fine tuning, should be restricted to a sample taken prior to the publication of
the target research.

\subsubsection{Group Testing}

\begin{figure}
\includegraphics[width=\textwidth]{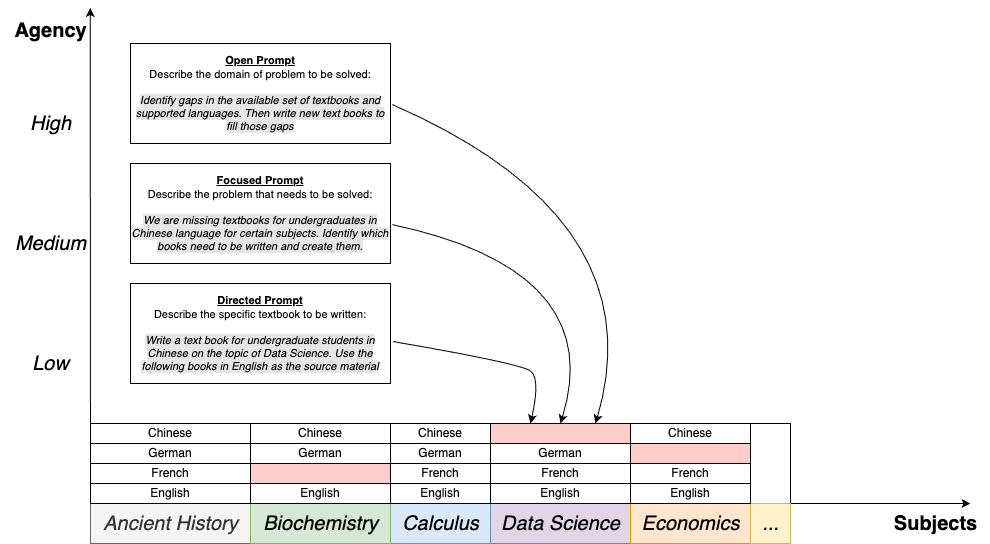}
\caption{Language cohort evaluation of AGI for education tasks.}
\label{fig:cohort}
\end{figure}

Group, cohort or cluster based testing in data science typically refers to methods
that seek to evaluate the ability of a model to generalise to out-of-sample data
belonging to specific groups. This technique forms a standard part of 
testing repertoires in libraries like SciKit Learn\cite{pedregosa2011scikit}, and
multiple variations of the approach have been presented to solve specific data science
problems\cite{Schneider2020,wecker-etal-2020-clusterdatasplit}. 
The general idea is to simulate some form of real world 
performance by creating test data that is novel across one or more dimensions. 
This approach is often applied
when it is known there are sampling biases in the training data, or that the
real world processes are changing such that distributions of future data are 
expected to be different from the training data.

This general idea can be applied to AGI evaluation to determine if the system is
able to generalise concepts across boundaries such as language, and recognise the
gaps in existing text corpora. We can apply this method to evaluate a system on 
its ability to transfer knowledge and 
insights across domains or even languages. We want to know that intelligence can fluidly take
knowledge and skills and apply them in a new domain with some degree of
novelty.

An example is shown in Figure \ref{fig:cohort},
involving multi-lingual text corpora with deliberate subject emissions. We want 
a system that at low agency can create a textbook for a subject that it has never
previously seen in the target language. While at high agency the system should be
able in independently recognise the
subjects that are missing in certain languages and fill these gaps. Such a system
would be capable of scanning literature and finding gaps in the market that are
worth addressing, even if the potential returns are low. Note, that as in the
out-of-time instance, the most important part of the evaluation protocol is ensuring
there is no leakage of the target output into the training data.

\subsubsection{Uncertainty Quantification}

A critical element of any data science solution is that the outputs of
a model can be calibrated such that costs and risks can be controlled. 
This typically means that model predictions are provided with a quantified 
level of uncertainty. Standard classification models can provide a probability
estimate for any prediction, allowing the system to use the probability
value when calibrating the actions that are taken. Other data science solutions,
such as forecasts of numerical quantities are provided uncertainty quantification
using methods such as conformal prediction\cite{angelopoulos2023conformal}.

Arguably, the inclusion of an estimate of uncertainty mimics human decision
making. Human workers tend to take decisive action only in situations of
reasonable certainty, 
they will delay or defer action under conditions of uncertainty, in a manner that is
sensitive to the nature of the uncertainty. Actions taken under
conditions of uncertainty can involve requests for additional information,
referring a case to a manager, or marking the task as requiring further review.

\begin{table}
\caption{Uncertainty Quantification in Administrative Tasks}\label{tab:admin}
\begin{tabular}{|l|l|l|}
\hline
Level             & Prompt\\
\hline
High              & Process the documents provided in each email in your inbox according to \\
                  & the rules in you job manual.\\
\hline
Medium            & Process the documents provided in each email in your inbox according to \\
                  & the rules in you job manual. Please pay particular attention to rules 7 \\
                  & and 42 for potential conflicts.\\
\hline
Low               & Process the documents provided in each email in your inbox according to \\
                  & the rules in you job manual. Please pay particular attention to rules 7 \\
                  & and 42 for potential conflicts. In some instances you will need to request \\
                  & additional information, in others the case will not be resolvable and must \\
                  & be referred to your manager. \\
\hline
\end{tabular}
\end{table}

To test potential AGI systems for adequate uncertainty quantification in their
reasoning we propose the development of an administrative task simulator with
controlled parameters. The task should consist of making decisions on the basis
of an open-ended number of documents according to a set of rules that define the
job. Some of these rules should be defined such that in a small subset of situations
the rules will conflict. Resolution of the conflicts either requires
request for additional information, or requires admission that the case cannot
be processed (hence marked for review). In Table \ref{tab:admin} we define a set of example
prompts that illustrate what agency might involve for such a task. Note that at low
agency the agent is told both where problems will emerge, and the set of potential
actions to take, while an agent with high autonomy will infer these operating principles
from the job manual itself.

\section{Conclusion}

Progress in any machine learning research endeavour depends heavily on strong methods
for evaluation. AGI is no exception to this rule, without robust and universal
methods of evaluation progress will be hampered by ongoing arguments about whether
results on a given benchmark truly represents general intelligence.

In this work we have argued for a pragmatic evaluation framework that focuses on
demonstrations of competence in real world tasks. These tasks should represent
the kinds of work that an AGI would be expected to do in an economy that
welcomes them as digital workers. We have drawn inspiration from proven
methods in data science, that have been used to successfully evaluate the 
performance of machine learning systems in real world tasks. These methods of
out-of-time testing, group analysis and uncertainty quantification encourage
robust decisions by simulating how a system will perform under conditions that
resemble the novelty and uncertainty of real-world tasks.

\begin{credits}

\subsubsection{\discintname}
The authors have no competing interests to declare that are
relevant to the content of this article. 
\end{credits}



\bibliographystyle{splncs04}
\bibliography{./refs}

\begin{thebibliography}{10}
\providecommand{\url}[1]{\texttt{#1}}
\providecommand{\urlprefix}{URL }
\providecommand{\doi}[1]{https://doi.org/#1}

\bibitem{Adams_Banavar_Campbell_2016}
Adams, S.S., Banavar, G., Campbell, M.: I-athlon: Towards a multidimensional
  turing test. AI Magazine  \textbf{37}(1),  78--84 (Apr 2016).
  \doi{10.1609/aimag.v37i1.2643},
  \url{https://ojs.aaai.org/aimagazine/index.php/aimagazine/article/view/2643}

\bibitem{Akman2000}
Akman, V., Blackburn, P.: Editorial: Alan turing and artificial intelligence.
  Journal of Logic Language and Information  \textbf{9} (10 2000).
  \doi{10.1023/A:1008389623883}

\bibitem{alomar2025}
AlOmar, E.A., DeMario, C., Shagawat, R., Kreiser, B.: Leakagedetector: An open
  source data leakage analysis tool in machine learning pipelines (2025),
  \url{https://arxiv.org/abs/2503.14723}

\bibitem{angelopoulos2023conformal}
Angelopoulos, A.N., Bates, S., et~al.: Conformal prediction: A gentle
  introduction. Foundations and Trends{\textregistered} in Machine Learning
  \textbf{16}(4),  494--591 (2023)

\bibitem{norvig2023}
y~Arcas, B.A., Norvig, P.: Artificial general intelligence is already here.
  Noema (2023),
  \url{https://www.noemamag.com/artificial-general-intelligence-is-already-here/}

\bibitem{10.1145/2783258.2783372}
Bifet, A., de~Francisci~Morales, G., Read, J., Holmes, G., Pfahringer, B.:
  Efficient online evaluation of big data stream classifiers. In: Proceedings
  of the 21th ACM SIGKDD International Conference on Knowledge Discovery and
  Data Mining. p. 59–68. KDD '15, Association for Computing Machinery, New
  York, NY, USA (2015). \doi{10.1145/2783258.2783372},
  \url{https://doi.org/10.1145/2783258.2783372}

\bibitem{blum2015}
Blum, A., Hardt, M.: The ladder: A reliable leaderboard for machine learning
  competitions (2015), \url{https://arxiv.org/abs/1502.04585}

\bibitem{Bringsjord2001}
Bringsjord, S., Bello, P., Ferrucci, D.: Creativity, the turing test, and the
  (better) lovelace test. Minds and Machines  \textbf{11},  3--27 (02 2001).
  \doi{10.1023/A:1011206622741}

\bibitem{bubeck2023sparksartificialgeneralintelligence}
Bubeck, S., Chandrasekaran, V., Eldan, R., Gehrke, J., Horvitz, E., Kamar, E.,
  Lee, P., Lee, Y.T., Li, Y., Lundberg, S., Nori, H., Palangi, H., Ribeiro,
  M.T., Zhang, Y.: Sparks of artificial general intelligence: Early experiments
  with gpt-4 (2023), \url{https://arxiv.org/abs/2303.12712}

\bibitem{Bugaj2009/06}
Bugaj, V., Goertzel, B.: Agi preschool: A framework for evaluating early-stage
  human-like agis. In: Proceedings of the 2nd Conference on Artificial General
  Intelligence (2009). pp. 12--17. Atlantis Press (2009/06).
  \doi{10.2991/agi.2009.3}, \url{https://doi.org/10.2991/agi.2009.3}

\bibitem{10.1145/3641289}
Chang, Y., Wang, X., Wang, J., Wu, Y., Yang, L., Zhu, K., Chen, H., Yi, X.,
  Wang, C., Wang, Y., Ye, W., Zhang, Y., Chang, Y., Yu, P.S., Yang, Q., Xie,
  X.: A survey on evaluation of large language models. ACM Trans. Intell. Syst.
  Technol.  \textbf{15}(3) (mar 2024). \doi{10.1145/3641289},
  \url{https://doi.org/10.1145/3641289}

\bibitem{chollet2019measureintelligence}
Chollet, F.: On the measure of intelligence (2019),
  \url{https://arxiv.org/abs/1911.01547}

\bibitem{Cohen2015}
Cohen, A., Smalheiser, N., McDonagh, M., Yu, C., Adams, C., Yu, P.: Automated
  confidence ranked classification of randomized controlled trial articles: An
  aid to evidence-based medicine. Journal of the American Medical Informatics
  Association : JAMIA  \textbf{22} (02 2015). \doi{10.1093/jamia/ocu025}

\bibitem{feldman2020}
Feldman, V.: Does learning require memorization? a short tale about a long
  tail. In: 52nd Annual ACM SIGACT Symposium on Theory of Computing. pp.
  954--959 (06 2020). \doi{10.1145/3357713.3384290}

\bibitem{goertzel2014artificial}
Goertzel, B.: Artificial general intelligence: concept, state of the art, and
  future prospects. Journal of Artificial General Intelligence  \textbf{5}(1),
  ~1 (2014)

\bibitem{guo2023evaluatinglargelanguagemodels}
Guo, Z., Jin, R., Liu, C., Huang, Y., Shi, D., Supryadi, Yu, L., Liu, Y., Li,
  J., Xiong, B., Xiong, D.: Evaluating large language models: A comprehensive
  survey (2023), \url{https://arxiv.org/abs/2310.19736}

\bibitem{hartmann2023}
Hartmann, V., Suri, A., Bindschaedler, V., Evans, D., Tople, S., West, R.: Sok:
  Memorization in general-purpose large language models (2023),
  \url{https://arxiv.org/abs/2310.18362}

\bibitem{10.1145/3676581.3676582}
Hawkins, J., Tivey, D.: Literature filtering for systematic reviews with
  transformers. In: Proceedings of the 2024 2nd International Conference on
  Communications, Computing and Artificial Intelligence. p. 1–7. CCCAI '24,
  Association for Computing Machinery, New York, NY, USA (2024).
  \doi{10.1145/3676581.3676582}, \url{https://doi.org/10.1145/3676581.3676582}

\bibitem{Hewamalage2022}
Hewamalage, H., Ackermann, K., Bergmeir, C.: Forecast evaluation for data
  scientists: common pitfalls and best practices. Data Mining and Knowledge
  Discovery  \textbf{37} (12 2022). \doi{10.1007/s10618-022-00894-5}

\bibitem{kaddour2023challengesapplicationslargelanguage}
Kaddour, J., Harris, J., Mozes, M., Bradley, H., Raileanu, R., McHardy, R.:
  Challenges and applications of large language models (2023),
  \url{https://arxiv.org/abs/2307.10169}

\bibitem{10.5555/1565455.1565458}
Legg, S., Hutter, M.: A collection of definitions of intelligence. In:
  Proceedings of the 2007 Conference on Advances in Artificial General
  Intelligence: Concepts, Architectures and Algorithms: Proceedings of the AGI
  Workshop 2006. p. 17–24. IOS Press, NLD (2007)

\bibitem{Levinson1991}
Levinson, R., Hsu, F.h., Marsland, T.A., Schaeffer, J., Wilkins, D.: The role
  of chess in artificial intelligence research. In: 12th international joint
  conference on Artificial intelligence. pp. 547--552. Morgan Kaufmann
  Publishers (01 1991)

\bibitem{Lokker2023}
Lokker, C., Bagheri, E., Abdelkader, W., Parrish, R., Afzal, M., Navarro-Ruan,
  T., Cotoi, C., Germini, F., Linkins, L., Haynes, b., Chu, L., Iorio, A.: Deep
  learning to refine the identification of high-quality clinical research
  articles from the biomedical literature: Performance evaluation. Journal of
  Biomedical Informatics  \textbf{142},  104384 (05 2023).
  \doi{10.1016/j.jbi.2023.104384}

\bibitem{10.1007/s10489-021-02735-2}
Maldonado, S., L\'{o}pez, J., Iturriaga, A.: Out-of-time cross-validation
  strategies for classification in the presence of dataset shift. Applied
  Intelligence  \textbf{52}(5),  5770–5783 (Mar 2022).
  \doi{10.1007/s10489-021-02735-2},
  \url{https://doi.org/10.1007/s10489-021-02735-2}

\bibitem{marcus2022}
Marcus, G.: Dear elon musk, here are five things you might want to consider
  about agi. Substack (2022),
  \url{https://garymarcus.substack.com/p/dear-elon-musk-here-are-five-things}

\bibitem{McCorduck2004}
McCorduck, P.: Machines Who Think: A Personal Inquiry into the History and
  Prospects of Artificial Intelligence. AK Peters Ltd (2004)

\bibitem{10.5555/3692070.3693548}
Morris, M.R., Sohl-Dickstein, J., Fiedel, N., Warkentin, T., Dafoe, A., Faust,
  A., Farabet, C., Legg, S.: Position: levels of agi for operationalizing
  progress on the path to agi. In: Proceedings of the 41st International
  Conference on Machine Learning. ICML'24, JMLR.org (2024)

\bibitem{Mueller2024}
Mueller, M.: The myth of agi: How the illusion of artificial general
  intelligence distorts and distracts digital governance (2024),
  \url{https://www.internetgovernance.org/wp-content/uploads/MythofAGI.pdf}

\bibitem{nakamoto2009bitcoin}
Nakamoto, S.: Bitcoin: A peer-to-peer electronic cash system  (May 2009),
  \url{http://www.bitcoin.org/bitcoin.pdf}

\bibitem{Nilsson_2005}
Nilsson, N.J.: Human-level artificial intelligence? be serious! AI Magazine
  \textbf{26}(4), ~68 (Dec 2005). \doi{10.1609/aimag.v26i4.1850},
  \url{https://ojs.aaai.org/aimagazine/index.php/aimagazine/article/view/1850}

\bibitem{pedregosa2011scikit}
Pedregosa, F., Varoquaux, G., Gramfort, A., Michel, V., Thirion, B., Grisel,
  O., Blondel, M., Prettenhofer, P., Weiss, R., Dubourg, V., et~al.:
  Scikit-learn: Machine learning in python. Journal of machine learning
  research  \textbf{12}(Oct),  2825--2830 (2011)

\bibitem{peng2024}
Peng, Y., Han, J., Zhang, Z., Fan, L., Liu, T., Qi, S., Feng, X., Ma, Y., Wang,
  Y., Zhu, S.C.: The tong test: Evaluating artificial general intelligence
  through dynamic embodied physical and social interactions. Engineering
  \textbf{34},  12--22 (2024). \doi{https://doi.org/10.1016/j.eng.2023.07.006},
  \url{https://www.sciencedirect.com/science/article/pii/S209580992300293X}

\bibitem{pfister2025}
Pfister, R., Jud, H.: Understanding and benchmarking artificial intelligence:
  Openai's o3 is not agi (2025), \url{https://arxiv.org/abs/2501.07458}

\bibitem{Schneider2020}
Schneider, S.A., Ohzeki, K., Geigis, M.: Leave a group out cross-validation
  (lagocv) to determine threshold for license plate detection in autonomus
  driving. International Journal of Computer Science and Applications
  \textbf{17},  15--32 (10 2020)

\bibitem{TASHMAN2000437}
Tashman, L.J.: Out-of-sample tests of forecasting accuracy: an analysis and
  review. International Journal of Forecasting  \textbf{16}(4),  437--450
  (2000). \doi{https://doi.org/10.1016/S0169-2070(00)00065-0},
  \url{https://www.sciencedirect.com/science/article/pii/S0169207000000650},
  the M3- Competition

\bibitem{tirumala2022}
Tirumala, K., Markosyan, A.H., Zettlemoyer, L., Aghajanyan, A.: Memorization
  without overfitting: Analyzing the training dynamics of large language models
  (2022), \url{https://arxiv.org/abs/2205.10770}

\bibitem{Vaswani2017}
Vaswani, A., Shazeer, N., Parmar, N., Uszkoreit, J., Jones, L., Gomez, A.N.,
  Kaiser, L.u., Polosukhin, I.: Attention is all you need. In: Guyon, I.,
  Luxburg, U.V., Bengio, S., Wallach, H., Fergus, R., Vishwanathan, S.,
  Garnett, R. (eds.) Advances in Neural Information Processing Systems.
  vol.~30. Curran Associates, Inc. (2017)

\bibitem{voss2007}
Voss, P.: Essentials of General Intelligence: The Direct Path to Artificial
  General Intelligence, pp. 131--157. Springer Berlin Heidelberg, Berlin,
  Heidelberg (2007). \doi{10.1007/978-3-540-68677-4_4}

\bibitem{Wang2010/06}
Wang, P.: The evaluation of agi systems. In: Proceedings of the 3d Conference
  on Artificial General Intelligence (2010). pp. 154--159. Atlantis Press
  (2010/06). \doi{10.2991/agi.2010.33},
  \url{https://doi.org/10.2991/agi.2010.33}

\bibitem{wang2019defining}
Wang, P.: On defining artificial intelligence. Journal of Artificial General
  Intelligence  \textbf{10}(2),  1--37 (2019)

\bibitem{wecker-etal-2020-clusterdatasplit}
Wecker, H., Friedrich, A., Adel, H.: {C}luster{D}ata{S}plit: Exploring
  challenging clustering-based data splits for model performance evaluation.
  In: Eger, S., Gao, Y., Peyrard, M., Zhao, W., Hovy, E. (eds.) Proceedings of
  the First Workshop on Evaluation and Comparison of NLP Systems. pp. 155--163.
  Association for Computational Linguistics, Online (Nov 2020).
  \doi{10.18653/v1/2020.eval4nlp-1.15},
  \url{https://aclanthology.org/2020.eval4nlp-1.15/}

\bibitem{Wozniak}
Wozniak, S.: Could a computer make a cup of coffee? (2010),
  \url{https://www.youtube.com/watch?v=MowergwQR5Y}

\bibitem{Xu2023}
Xu, B., Ren, Q.: Artificial open world for evaluating agi: A conceptual
  design. In: Goertzel, B., Ikl{\'e}, M., Potapov, A., Ponomaryov, D. (eds.)
  Artificial General Intelligence. pp. 452--463. Springer International
  Publishing, Cham (2023)

\bibitem{ye2024}
Ye, J., Borovykh, A., Hayou, S., Shokri, R.: Leave-one-out distinguishability
  in machine learning (2024), \url{https://arxiv.org/abs/2309.17310}

\bibitem{zhong2024}
Zhong, T., Liu, Z., Pan, Y., Zhang, Y., Zhou, Y., Liang, S., Wu, Z., Lyu, Y.,
  Shu, P., Yu, X., Cao, C., Jiang, H., Chen, H., Li, Y., Chen, J., Hu, H., Liu,
  Y., Zhao, H., Xu, S., Dai, H., Zhao, L., Zhang, R., Zhao, W., Yang, Z., Chen,
  J., Wang, P., Ruan, W., Wang, H., Zhao, H., Zhang, J., Ren, Y., Qin, S.,
  Chen, T., Li, J., Zidan, A.H., Jahin, A., Chen, M., Xia, S., Holmes, J.,
  Zhuang, Y., Wang, J., Xu, B., Xia, W., Yu, J., Tang, K., Yang, Y., Sun, B.,
  Yang, T., Lu, G., Wang, X., Chai, L., Li, H., Lu, J., Sun, L., Zhang, X., Ge,
  B., Hu, X., Zhang, L., Zhou, H., Zhang, L., Zhang, S., Liu, N., Jiang, B.,
  Kong, L., Xiang, Z., Ren, Y., Liu, J., Jiang, X., Bao, Y., Zhang, W., Li, X.,
  Li, G., Liu, W., Shen, D., Sikora, A., Zhai, X., Zhu, D., Liu, T.: Evaluation
  of openai o1: Opportunities and challenges of agi (2024),
  \url{https://arxiv.org/abs/2409.18486}

\end{thebibliography}

\end{document}